\documentclass[letterpaper, 10 pt, conference]{ieeeconf} 
\IEEEoverridecommandlockouts
\usepackage[hidelinks]{hyperref}
\hypersetup{breaklinks=true}
\usepackage{fix-cm}
\usepackage{etex}

\usepackage{dblfloatfix}

\usepackage{nag}

\makeatletter
\@ifpackageloaded{xcolor}{}{%
\usepackage[table,x11names,dvipsnames,svgnames]{xcolor}%
}
\makeatother

\usepackage{colortbl}

\usepackage{graphicx}
\usepackage{wrapfig}

\definecolorset{RGB}{lyft}{}{Red,194,39,36;Sunset,202,53,33;Orange,205,68,20;Amber,200,117,42;Yellow,242,169,52;Citron,186,188,44;Lime,112,159,33;Green,56,139,31;Mint,45,118,56;Teal,52,133,135;Cyan,60,132,202;Blue,55,94,248;Indigo,64,13,247;Purple,115,42,248;Pink,176,25,145;Rose,176,32,75}

\usepackage{cite}

\usepackage{microtype}

\usepackage[american]{babel}

\usepackage{array}
\usepackage{multirow}
\usepackage{booktabs}
\usepackage{makecell} 

\ifcsname labelindent\endcsname
\let\labelindent\relax
\fi
\usepackage[inline]{enumitem}

\usepackage{subfig}

\newenvironment{lenumerate}[2][]
{\begin{enumerate}[label=#2\arabic*:,leftmargin=0.2in,itemindent=0.15in,#1]}
{\end{enumerate}}

\setlist*[enumerate,1]{label={\itshape\arabic*)}}

\makeatletter
\newcommand{\paragraphswithstop}{%
\let\copyparagraph\paragraph%
\renewcommand\paragraph[1]{\copyparagraph{##1.}}%
}
\makeatother

\usepackage[framemethod=tikz]{mdframed}

\makeatletter
\def\namedlabel#1#2{\begingroup
  #2%
  \def\@currentlabel{#2}%
  \phantomsection\label{#1}\endgroup
}
\makeatother

\usepackage{suffix}

\usepackage{environ}

\makeatletter
\newsavebox{\boxifnotempty}
\newcommand{\displayifnotempty}[3]{\sbox\boxifnotempty{#2}\setbox0=\hbox{\usebox{\boxifnotempty}\unskip}%
\ifdim\wd0=0pt
\else
 #1\usebox{\boxifnotempty}#3%
\fi%
}
\newcommand{\ifempty}[2]{\setbox0=\hbox{#1\unskip}%
\ifdim\wd0=0pt%
 #2%
\fi%
}
\newcommand{\ifnotempty}[2]{\setbox0=\hbox{#1\unskip}%
\ifdim\wd0>0pt%
 #2%
\fi%
}
\makeatother

\usepackage{algorithm}

\usepackage{scrlfile}

\makeatletter
\newcommand*\newstoreddef[1]{
  \BeforeClosingMainAux{%
    \immediate\write\@auxout{%
      \string\restoredef{#1}{\csname #1\endcsname}%
    }%
  }%
}
\newcommand*{\restoredef}[2]{
  \expandafter\gdef\csname stored@#1\endcsname{#2}%
}
\newcommand*{\storeddef}[1]{
  \@ifundefined{stored@#1}{0}{\csname stored@#1\endcsname}%
}
\makeatother

\usepackage{pageslts}
\NewEnviron{tee}{\BODY\typeout{Marker Tee [start] ^^J \BODY ^^JMaker Tee [end]}}

\input{preamble/math}

\newcommand{\bmat}[1]{\begin{bmatrix}#1\end{bmatrix}}

\DeclarePairedDelimiter{\norm}{\lVert}{\rVert}

\DeclareMathOperator*{\argmin}{\arg\!\min}

\usepackage{units}

  \newcommand{\newcolorlabel}[2]{%
  \expandafter\newcommand\csname #1\endcsname[1]{%
    \tikz[baseline]{\node[text=white,fill=#2,anchor=base,text height=1.3ex,text depth=0.1ex,font=\sffamily\bfseries]{##1}}}%
}

\newcommand{\newcommenter}[2]{%
  \expandafter\newcommand\csname #1\endcsname[1]{%
    \fcolorbox{#2}{#2}{\color{white}\textsf{\textbf{#1}}}
    {\color{#2}##1}}%
  \expandafter\newcommand\csname at#1\endcsname{%
    \fcolorbox{#2}{#2}{\color{white}\textsf{\textbf{@#1}}}
    {\color{#2}}}%
  \expandafter\newcommand\csname #1hl\endcsname[2]{%
    \colorbox{#2}{\color{white}\textsf{\textbf{#1}}}\sethlcolor{Azure2}\hl{##2}~%
    \expandafter\ifx\csname commentarrow\endcsname\relax$\leftarrow$\else \commentarrow[#2]\fi~%
    {\color{#2}##1}}%
  \expandafter\newcommand\csname #1st\endcsname[2]{%
    \colorbox{#2}{\color{white}\textsf{\textbf{#1}}}\sout{##2}~%
    \expandafter\ifx\csname commentarrow\endcsname\relax$\leftarrow$\else \commentarrow[#2]\fi~%
    {\color{#2}##1}}%
}
\newcommenter{TODO}{DodgerBlue1}
\newcommenter{rtron}{Green3}

\usepackage{comment}

\usepackage{pdfcomment}

\usepackage{soul}

\usepackage[normalem]{ulem}

\usepackage{csquotes}

\usepackage{tikz}
\usetikzlibrary{calc}
\usetikzlibrary{matrix}
\usetikzlibrary{chains}
\usetikzlibrary{shapes.geometric}
\usetikzlibrary{arrows.meta}
\usetikzlibrary{decorations.pathreplacing}
\usetikzlibrary{backgrounds}

\tikzset{
  dim above/.style={to path={\pgfextra{
        \pgfinterruptpath
        \draw[>=latex,|->|] let
        \p1=($(\tikztostart)!1.5em!90:(\tikztotarget)$),
        \p2=($(\tikztotarget)!1.5em!-90:(\tikztostart)$)
        in(\p1) -- (\p2) node[pos=.5,sloped,above]{#1};
        \endpgfinterruptpath
      }
    }
  },
  dim double above/.style={to path={\pgfextra{
        \pgfinterruptpath
        \draw[>=latex,|->|] let
        \p1=($(\tikztostart)!3em!90:(\tikztotarget)$),
        \p2=($(\tikztotarget)!3em!-90:(\tikztostart)$)
        in(\p1) -- (\p2) node[pos=.5,sloped,above]{#1};
        \endpgfinterruptpath
      }
    }
  },
  dim below/.style={to path={\pgfextra{
        \pgfinterruptpath
        \draw[>=latex,|->|] let 
        \p1=($(\tikztostart)!-1em!-90:(\tikztotarget)$),
        \p2=($(\tikztotarget)!-1em!90:(\tikztostart)$)
        in (\p1) -- (\p2) node[pos=.5,sloped,below]{#1};
        \endpgfinterruptpath
      }
    }
  },
}

\tikzset{
    right angle quadrant/.code={
        \pgfmathsetmacro\quadranta{{1,1,-1,-1}[#1-1]}     
        \pgfmathsetmacro\quadrantb{{1,-1,-1,1}[#1-1]}},
    right angle quadrant=1, 
    right angle length/.code={\def\rightanglelength{#1}},   
    right angle length=2ex, 
    right angle symbol/.style n args={3}{
        insert path={
            let \p0 = ($(#1)!(#3)!(#2)$) in     
                let \p1 = ($(\p0)!\quadranta*\rightanglelength!(#3)$), 
                \p2 = ($(\p0)!\quadrantb*\rightanglelength!(#2)$) in 
                let \p3 = ($(\p1)+(\p2)-(\p0)$) in  
            (\p1) -- (\p3) -- (\p2)
        }
    }
}

\newcommand{\pgfextractangle}[3]{%
    \pgfmathanglebetweenpoints{\pgfpointanchor{#2}{center}}
                              {\pgfpointanchor{#3}{center}}
    \global\let#1\pgfmathresult  
}

\usetikzlibrary{shapes.arrows}
\newcommand{\commentarrow}[1][Azure4]{\tikz[baseline=-3pt]{\node[shape border uses incircle, fill=#1,rotate=180,single arrow, inner sep=1pt, minimum size=6pt, single arrow head extend=2pt]{};}}

\tikzset{ax/.style={-latex,line width=2pt}}

\tikzset{camera/.style={fill=Sienna1,fill opacity=0.5},%
image plane/.style={draw=RoyalBlue3,line width=2pt}}

\newcommenter{dawei}{Blue}
\newcommenter{rebecca}{Red}
\newcommenter{style}{olive}
\newcommenter{delete}{orange}
\usepackage{amsmath}

\newtheorem{hypothesis}{Hypothesis}
\title{\LARGE \bf
Haptic Feedback Improves Human-Robot Agreement and User Satisfaction in Shared-Autonomy Teleoperation
}

\author{Dawei Zhang$^{1}$, Roberto Tron$^{2}$, and Rebecca P. Khurshid$^{2}$
\thanks{$^{1}$Dawei Zhang is with the Department of Mechanical Engineering, Boston University, Boston, MA 02215, USA
        {\tt\small dwzhang@bu.edu}}%
\thanks{$^{2}$Roberto Tron and Rebecca P. Khurshid are with the Department of Mechanical Engineering 
8
 and the Division of Systems Engineering, Boston University, Boston, MA 02215, USA
        {\tt\small tron@bu.edu khurshid@bu.edu}}%
}

\begin{document}
\maketitle
 \begin{abstract}
Shared autonomy teleoperation can guarantee safety, but does so by reducing the human operator's control authority, which can lead to reduced levels of human-robot agreement and user satisfaction. This paper presents a novel haptic shared autonomy teleoperation paradigm that uses haptic feedback to inform the user about the inner state of a shared autonomy paradigm, while still guaranteeing safety. This differs from haptic shared control, which uses haptic feedback to inform the user's actions, but gives the human operator full control over the robot's actions. 
We conducted a user study in which twelve users flew a simulated UAV in a search-and-rescue task with no assistance or assistance provided by haptic shared control, shared autonomy, or haptic shared autonomy. All assistive teleoperation methods use control barrier functions to find a control command that is both safe and as close as possible to the human-generated control command. For assistive teleoperation conditions with haptic feedback, we apply a force to the user that is proportional to the difference between the human-generated control and the safe control. We find that haptic shared autonomy improves the user's task performance and satisfaction. We also find that haptic feedback in assistive teleoperation can improve the user's situational awareness. Finally, results show that adding haptic feedback to shared-autonomy teleoperation can improve human-robot agreement. 
\end{abstract}

\section{INTRODUCTION}\label{sec:intro}

\begin{figure}[t] 
  \centering
  \vspace{-10pt}
  \subfloat[No Assistance (NA) \label{fig:NA}]{
    \includegraphics[width=0.8\columnwidth, trim={0cm 0.2cm 0 0cm},clip]{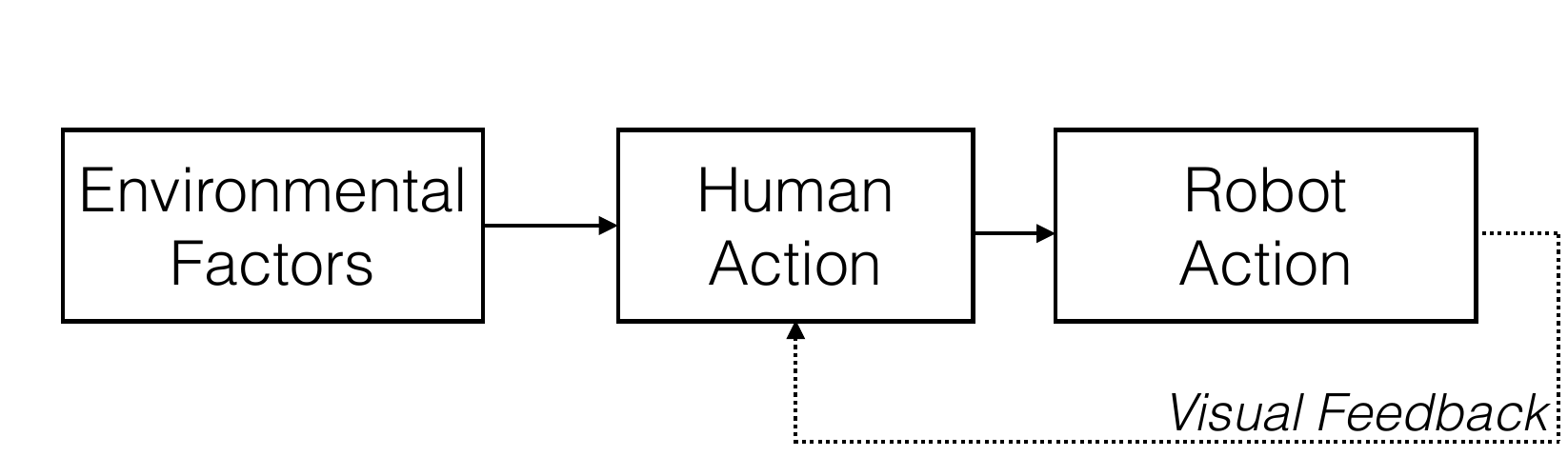}
    }
  \vspace{-8pt}  
  \subfloat[Haptic Shared Control (HSC) \label{fig:HSC}]{
    \includegraphics[width=0.8\columnwidth, trim={0cm 0.2cm 0 0cm},clip]{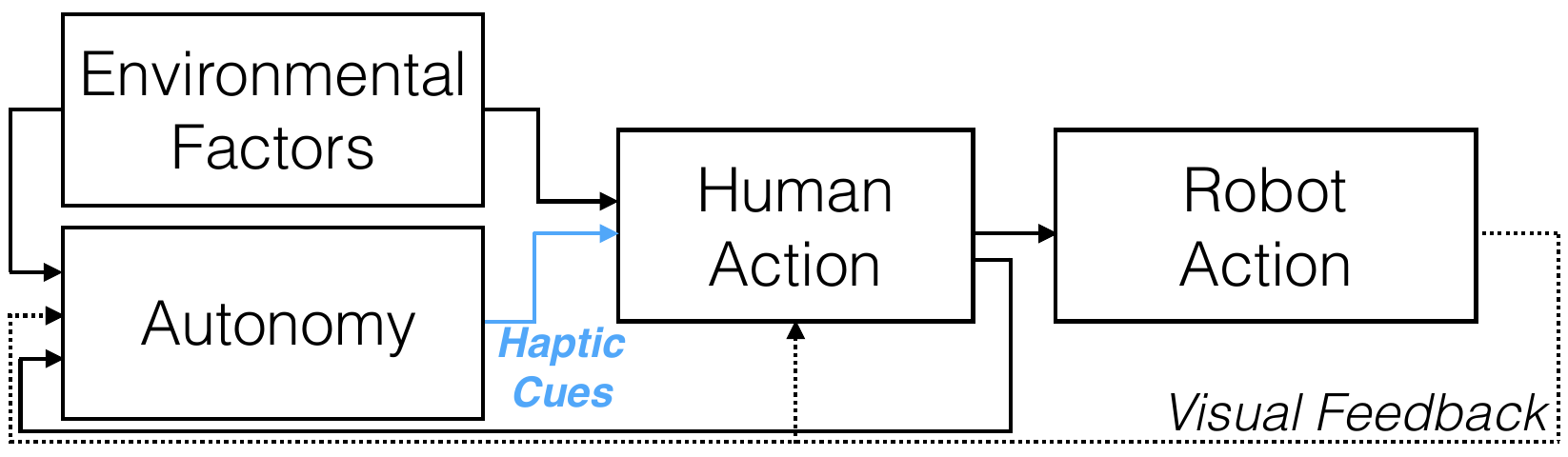}
  }
  \vspace{-10pt}
  \subfloat[Shared Autonomy (SA) \label{fig:SA}]{
    \includegraphics[width=0.8\columnwidth, trim={0cm 0.2cm 0 0cm},clip]{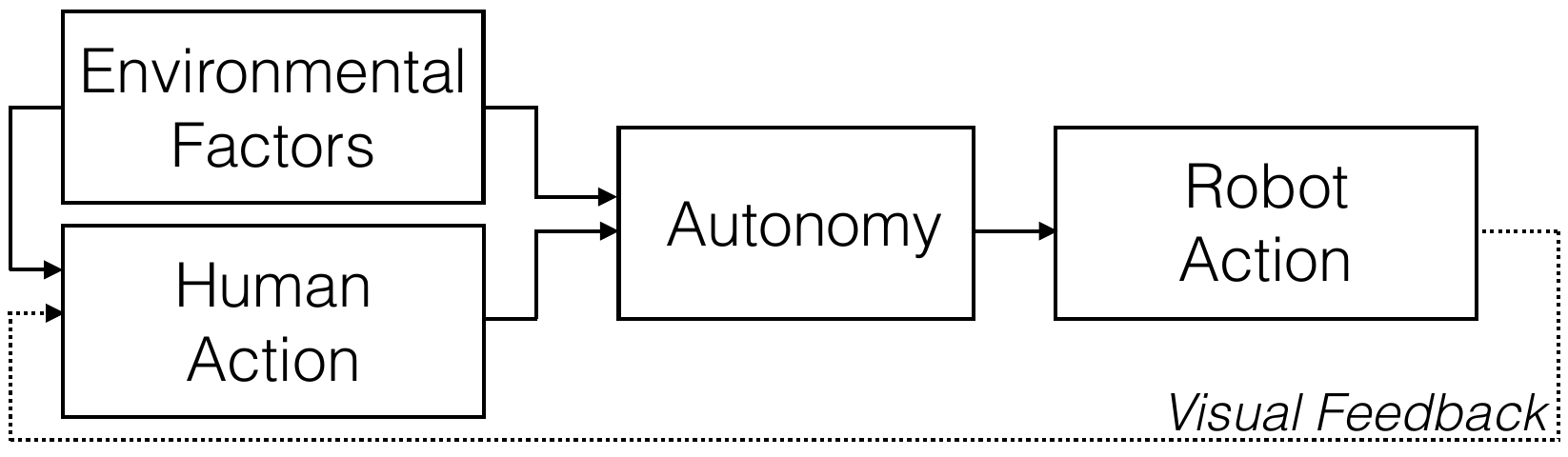}
  }
  \vspace{-10pt}
  \subfloat[Haptic Shared Autonomy (HSA) \label{fig:HSA}]{
    \includegraphics[width=0.8\columnwidth, trim={0cm 0.2cm 0 0cm},clip]{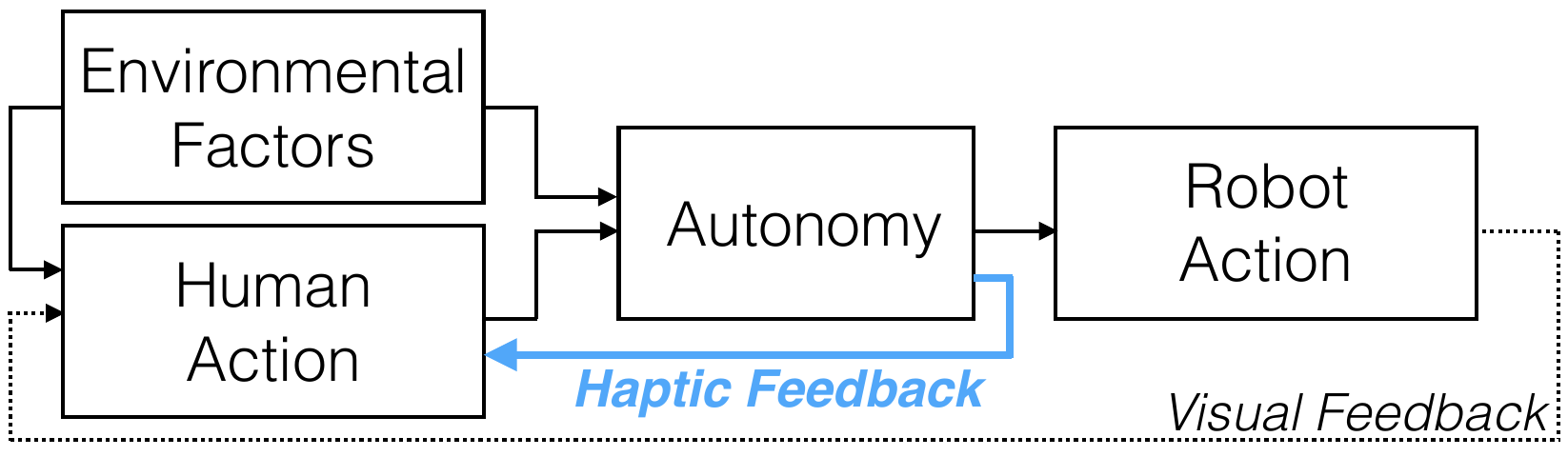}
  }
  \caption{\small Paradigms of shared control and autonomy for teleoperation with and without assistance.}
  \label{fig:paradigms}
  \vspace{-15pt}
\end{figure}

Teleoperation plays an important role in the robotics field by allowing humans to remotely work in hard-to-reach or hazardous environments. In traditional teleoperation systems (see Figure \ref{fig:NA}), the human operator has full control over all of the robot's actions, and \textbf{no assistance (NA)} is provided by the teleoperation system \cite{Niemeyer2008}. Unfortunately, it can be very challenging for a human operator to effectively use such a teleoperation system. For example, when flying a UAV, the human operator typically receives a limited field of view, which often leads to low levels of situational awareness, making it difficult to safely and accurately control the UAV \cite{mccarley2005human, Brandt2010}. Therefore, substantial research has been dedicated to creating systems that help the human operator complete their desired task with higher levels of efficiency and safety.  

\textbf{Haptic shared control (HSC)} has proven to be a promising approach to help the human operator remotely control a robot by providing haptic cues to the operator \cite{okamura2009haptic,pacchierotti2015cutaneous}. As illustrated in Figure \ref{fig:HSC}, the human will feel these haptic cues but will still maintain full control authority over the robot. For example, in a driving task, a haptic shared controller might apply a torque to a steering wheel to help keep that car in its current lane \cite{abbink2012haptic}. The human operator can choose to follow the haptic suggestions, or they can overpower their forces to override them. 
The fact that the human operator maintains full control authority over the robot is both a strength and weakness of haptic shared control systems, depending on the application. In the case of remote control of agile UAVs, it has been repeatedly shown that although haptic shared control can improve a user's ability to fly the UAV, human operators still tend to crash UAVs even when using haptic shared control \cite{Lam2009,Brandt2010, zhang2020haptic}.


\textbf{Shared autonomy (SA)} is an alternative approach to improve the robotic teleoperation by providing some levels of autonomous assistance. As shown in Fig. \ref{fig:SA}, the direct link between the human's control command and the robot's action is broken by the autonomous controller. Therefore, in shared autonomy teleoperation, the human no longer maintains full control authority of the robot's immediate action. Therefore, it can be difficult for the human operator to understand how and why their commands are modified by the autonomous assistance, which can lead to low levels of inner compatibility between the human and the shared autonomy system. The reduced transparency of the internal control loop can reduce users' satisfaction and their willingness to use the robot~\cite{javdani2018shared, pocius2020communicating}.\linebreak 
To address the challenge of balancing the task performance and safety, with system transparency and user satisfaction,
we propose a \textbf{haptic shared autonomy (HSA)} teleoperation paradigm. As shown in Fig. \ref{fig:HSA}, haptic shared autonomy uses haptic feedback to help inform the user about the inner state of a shared autonomy system. Specifically, we propose that haptic feedback can better enable the human operator to understand how their commands compare to the action ultimately taken by the robot. We propose that this will lead to improved agreement between the human operator and the autonomous assistance, and also improve operator satisfaction. 


In synthesis, the main contribution of this paper lies in proposing and evaluating a novel haptic shared autonomy teleoperation paradigm that takes the advantage of both shared autonomy and the ability to communicate safe actions via haptics feedback.


\section{RELATED WORK}\label{sec:related_work}
\subsection{Haptic Shared Control}
One example of haptic shared control is in the use of virtual fixtures, which have been widely implemented to generate force feedback when the human operator commands the robot to an area protected by the virtual fixture \cite{abbott2007haptic,bowyer2013active}. Other haptic shared control systems provide the human with warnings of a risk of collision between the robot and some protected region of space. For example,  Lam et al. proposed a parametric risk field (PRF) to calculate the risk of a collision \cite{Lam2009}.  Brant and Colton set the magnitude of the haptic feedback to be proportional to the time that it would take the UAV to collide with an object in its environment \cite{Brandt2010}. Recently, our prior work introduced a haptic shared control system that used control barrier functions to help guide the human towards the input command that is closed to their current command and deemed to be safe, rather than simply warning them about the risk of collision \cite{zhang2020haptic}. Haptic shared control has also been implemented to help a human operator perform grasping tasks with a robotic arm. 
Abi-Farraj et al. used haptic feedback to help navigate a robotic arm to a predicted grasping pose among multiple targets \cite{abi2019haptic}.
In a similar setup, Pocius et al. also proposed to use forces to communicate the shared control's prediction of the human's goal to the human operator \cite{pocius2020communicating}.
\subsection{Shared Autonomy}
Like haptic shared control, shared autonomy has been used to help improve safety for a human-controlled robot. Because the shared autonomy will command the robot's final action, shared autonomy systems can guarantee safety. 
Recently, Xu and Sreenath used Control Barrier Functions (CBFs) as a supervisory controller that modifies the human operator's control input and guarantees safe teleoperation of UAVs \cite{Xu2018}. Similar shared-autonomy paradigms include outer-loop stabilization \cite{broad2018learning} and parallel autonomy \cite{schwarting2017parallel}.  For example, Schwarting et al. applied  Non-linear Model Predictive Control (NMPC) to guarantee the safety of human-controlled automated vehicles \cite{schwarting2017parallel}. Shared autonomy has also been applied to assistive systems that predict the user’s intent, then alter the user's commands so that the robot makes better progress towards achieving the predicted human's goal \cite{Dragan2013,javdani2018shared,reddy2018shared}.
\vspace{-5pt}
\subsection{Haptic Shared Autonomy}
Although haptic feedback has been widely used for haptic shared control, haptic feedback has not been well-studied in the context of shared autonomy teleoperation. Masone et al. implemented haptic feedback in a shared planning method, in which the robot navigated a path whose parameters were jointly controlled by the human operator and an autonomous algorithm that ensured the path was safe and suitable for the task \cite{masone2018shared}. The team used haptic feedback to reflect the difference between the planned and traveled path. A user study found that the use of haptic feedback made it easier for the user to manipulate the robot's planned path. The effect of haptic feedback has also been studied in a shared autonomy paradigm, where the autonomous assistance fully takes over some aspect of the task and disregards the human's input. Griffin et al. implemented a shared autonomy system to help a human better control a robot manipulator. In this system, the autonomous assistance would regulate the manipulator's grip force so as to not drop an object in the robot's gripper. The authors were found to reflect the operator's commanded grip, rather than the force measured at the robot's fingertips \cite{griffin2005feedback}. To our knowledge haptic feedback has not been used in shared autonomy teleoperation when the human and autonomous assistance are working intimately together to have an immediate effect on the same aspect of the robot's action, for example the robot's position or velocity.
\section{METHODOLOGY} \label{sec:methods}
As discussed in Section \ref{sec:intro}, there are many approaches that could be taken to create assistive teleoperation paradigms. In this research study, we use control barrier functions (CBFs) to find the control signal that is as close as possible to the human commanded control input and would also guarantee that the UAV does not collide with an obstacle. In this section, we first discuss our model of the UAV's dynamics. We then briefly introduce CBFs with respect to our second order system. More details about CBFs can be found in \cite{Ames2014,Nguyen2016,Xiao2019}.

\subsection{Quadrotor UAV Dynamics and Control}

In this paper, we consider quadrotor UAVs that fly at relatively low speeds without highly aggressive maneuvers, so that the roll and pitch angles of the quadrotor will remain small. Under such conditions, the dynamics of the UAV can be modeled by a double integrator, where the control input $u$ corresponds to the acceleration command of the UAV. Let $x=(x_1,x_2)$ with $x_1$ being the position of the UAV, and $x_2 = \dot{x}_1$ its velocity. The dynamics of the system then become:

\begin{equation}\label{eqn:dynamics}
  \bmat{\dot{x}_{1} \\
    \dot{x}_{2}}=\bmat{
    0 & 1 \\
    0 & 0}
  \bmat{
    x_{1} \\
    x_{2}}
  +\bmat{
    0 \\
    1} u.
\end{equation}

We can write \eqref{eqn:dynamics} in a matrix form as:
\begin{equation}\label{eqn:dynamics_matrix}
    \dot{x}=Ax+Bu,
\end{equation}
where $A \in \mathbb{R}^{n \times n}$, $B \in \mathbb{R}^{n \times m}$ and $u \in \mathbb{R}^{m}$.
We implement a rate-control scheme, in which the position of the control interface, $p_i$, is scaled by a constant factor, $K_{v}$, to generate a desired velocity command, $x_{2,d} =  K_{v}p_i $ for the UAV. 
We then define the user's reference control input as: 
\begin{equation} 
u_{ref}=\frac{1}{\Delta t}(x_{2d}-x_2),
\end{equation}
where $x_2$ is the velocity of the UAV and $\Delta t$ is the rate of the control loop.

\subsection{Control Barrier Functions}\label{subsec:CBF}
\subsubsection{Notation}
We denote the lie derivative of a function $h(x)$ along the dynamics \eqref{eqn:dynamics_matrix} as $L_{f} h(x):=\frac{\partial h(x(t))}{\partial x(t)} Ax$,  $L_{g} h(x):=\frac{\partial h(x(t))}{\partial x(t)} B$. We use $L_{f}^{b} h(x)$ to denote a Lie derivative of order $b$.
\subsubsection{Safety Set}
A continuously differentiable function $h(x)$ can be define a  safety set $\mathcal{C}$, as follows:
\begin{equation}\label{eqn:safety_set}
\mathcal{C} : = \left\{ x \in \mathbb { R } ^ { n } : h (x )\geq 0 \right\}. 
\end{equation}


\subsubsection{CBFs for Second Order Systems}
The goal of control barrier functions is to always produce a control signal so that if $x_0 \in \mathcal{C} $ implies $x(t) \in \mathcal{C}, \forall t \in \mathrm{I}\left(x_{0}\right) $, i.e. the system is forward invariant \cite{Ames2019}. For a second order control system, a second-order exponential control barrier function can be used to constrain $u$ to ensure safe control. That is that if the state of system is in the safe set, $\mathcal{C}$, then any control signal $u$ from the set: 
\begin{multline}
U_{CBF} = \{ u \in U : L_{f}^{2} h(x)+L_{g} L_{f} h(x) u \\
+K\bmat{h(x)& L_{f}h(x)}^T \geq 0 \}, 
\end{multline}
will result in the state of the system remaining in the safety set \cite{Nguyen2016} when $K$ is chosen such that the real parts of eigenvalues of matrix $A-BK$ are negative \cite{Nguyen2016}.


\subsection{Assistive Teleoperation via CBFs}

We can use a Quadratic Program with CBF constraints to find a control input that is as close as possible to the human's commanded control while also guaranteeing safety. This formulation results in the following optimization problem:
\begin{equation}\label{eq:CBF-QP}
    \begin{array} { ll }
    {u_{CBF} =\underset{u \in \mathbb{R}^{m}}{\argmin}} {\frac{1}{2}\norm{u -u_{ref}}^{2}}\\
    \text { s.t. }L_{f}^{2} h(x)+L_{g} L_{f} h(x) u+K\bmat{h(x)& L_{f}h(x)}^T\geq 0,
    \end{array}
\end{equation}
where $u_{ref}$ is the reference control input that is provided by the human operator and $u_{CBF}$ is the safe control input. As summarized in Table \ref{table:conditions}, we can use $u_{ref}$ and $u_{CBF}$ to create a teleoperation condition that offers no assistance (NA), as well as haptic shared control (HSC), shared autonomy (SA), and haptic shared autonomy (HSA) teleoperation. In NA and HSC, the humans control input $u_{ref}$ is sent to the UAV, giving the human full control authority. In SA and HSA, the safe control command $u_{CBF}$ is used. HSC and HSA both use force feedback to inform the operator about the difference between their command and the closest safe control:
\begin{equation} 
F = K_f(u_{CBF}-u_{ref}),
\end{equation}
where $K_{f}$ is a constant parameter to adjust the magnitude of the haptic feedback.

\begin{table}[]
\caption{Teleoperation Conditions} \label{table:conditions}

\centering
\resizebox{\columnwidth}{!}{%
\begin{tabular}{@{}|l|l|l|@{}}
\hline
Condition & Control Signal & Haptic Feedback              \\ \hline 
NA        & $u_{ref}$                  & None                        \\
HSC       & $u_{ref}$                  & $K_f(u_{CBF}-u_{ref})$                        \\
SA        & $u_{CBF}$                  & None      \\
HSA       & $u_{CBF}$                  & $K_f(u_{CBF}-u_{ref})$ \\
\hline
\end{tabular}%
}\vspace{-15pt}
\end{table}

\section{USER STUDY DESIGN}\label{sec:user_study}
The goal of this paper is to assess whether haptic feedback can improve the user experience in a shared autonomy setting. Therefore, we designed a user study to test the following: 
\begin{hypothesis}\label{H1}
  HSA improves the user's task performance (speed and safety) and satisfaction in teleoperation tasks with respect to other paradigms shown in Table \ref{table:conditions}.
\end{hypothesis}

During the user study, participants completed a simulated search-and-rescue task by using a simulated UAV to locate targets of interest in a forest-like environment under each of the four conditions (NA, HSC, SA, and HSA). We designed the user study to also enable us to test the following specific hypotheses, which together support our main hypothesis. 


\begin{hypothesis}\label{H2}
   Teleoperation paradigms with haptic feedback (HSC and HSA) will result in better situational awareness compared with other paradigms (NA and SA). 
\end{hypothesis}
\begin{hypothesis}\label{H3}
    Teleoperation paradigms that include shared autonomy (SA and HSA) will result in better task performance compared with other paradigms (NA and HSC).
\end{hypothesis}
\begin{hypothesis}\label{H4}
   HSA will lead to better human-robot agreement as compared with SA.
\end{hypothesis}

The study was approved by the Boston University Institutional Review Board under protocol number 5070E.

\subsection{Experimental Setup}
\begin{figure}[t]
  \centering
  \includegraphics[width=0.9\columnwidth, trim={8cm 0cm 27cm 3cm},clip]{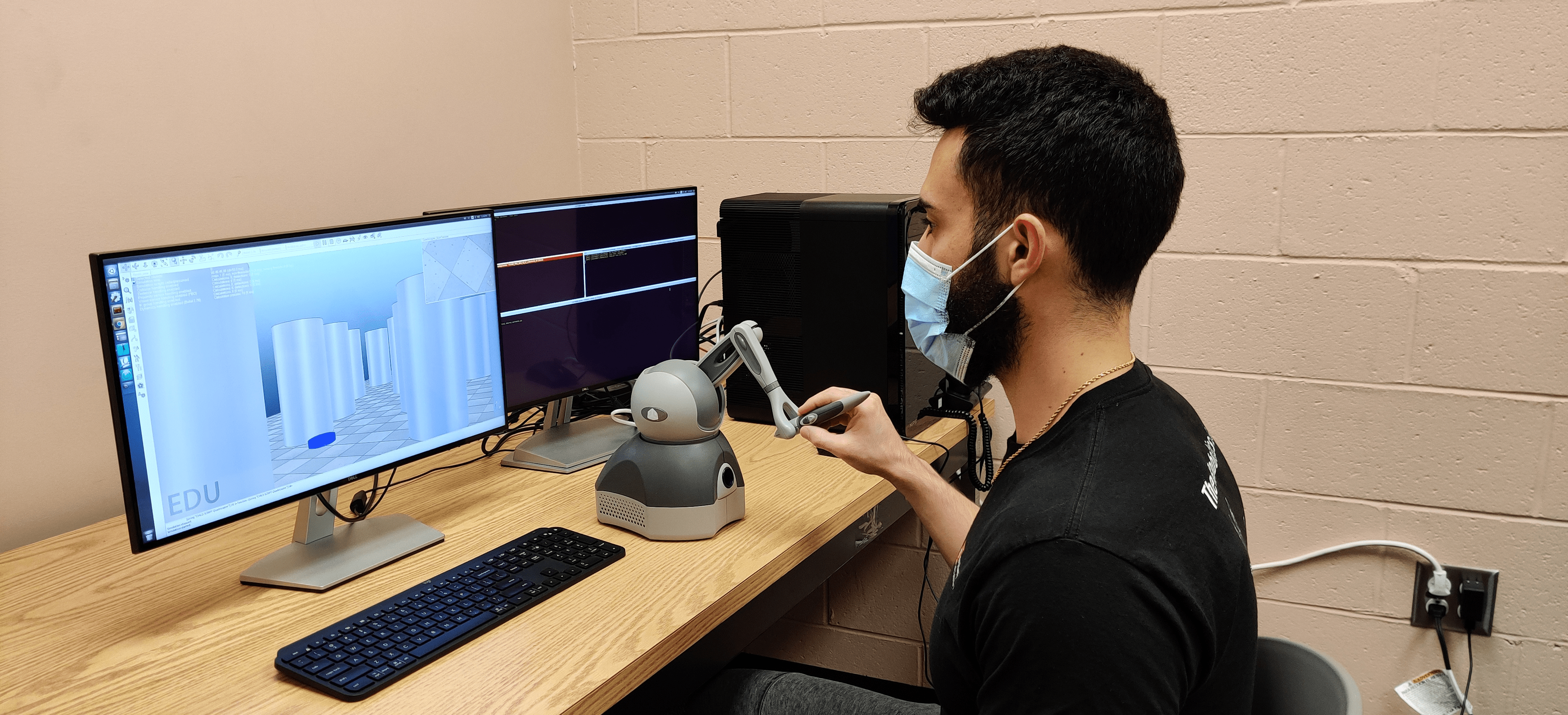}
  \caption{\small Each participant used a haptic joystick to control the UAV in a simulated environment.}
  \label{fig:setup}
\end{figure}

\begin{figure}[t]
  \centering
  \includegraphics[width=0.9\columnwidth, trim={0cm 0cm 0cm 0cm},clip]{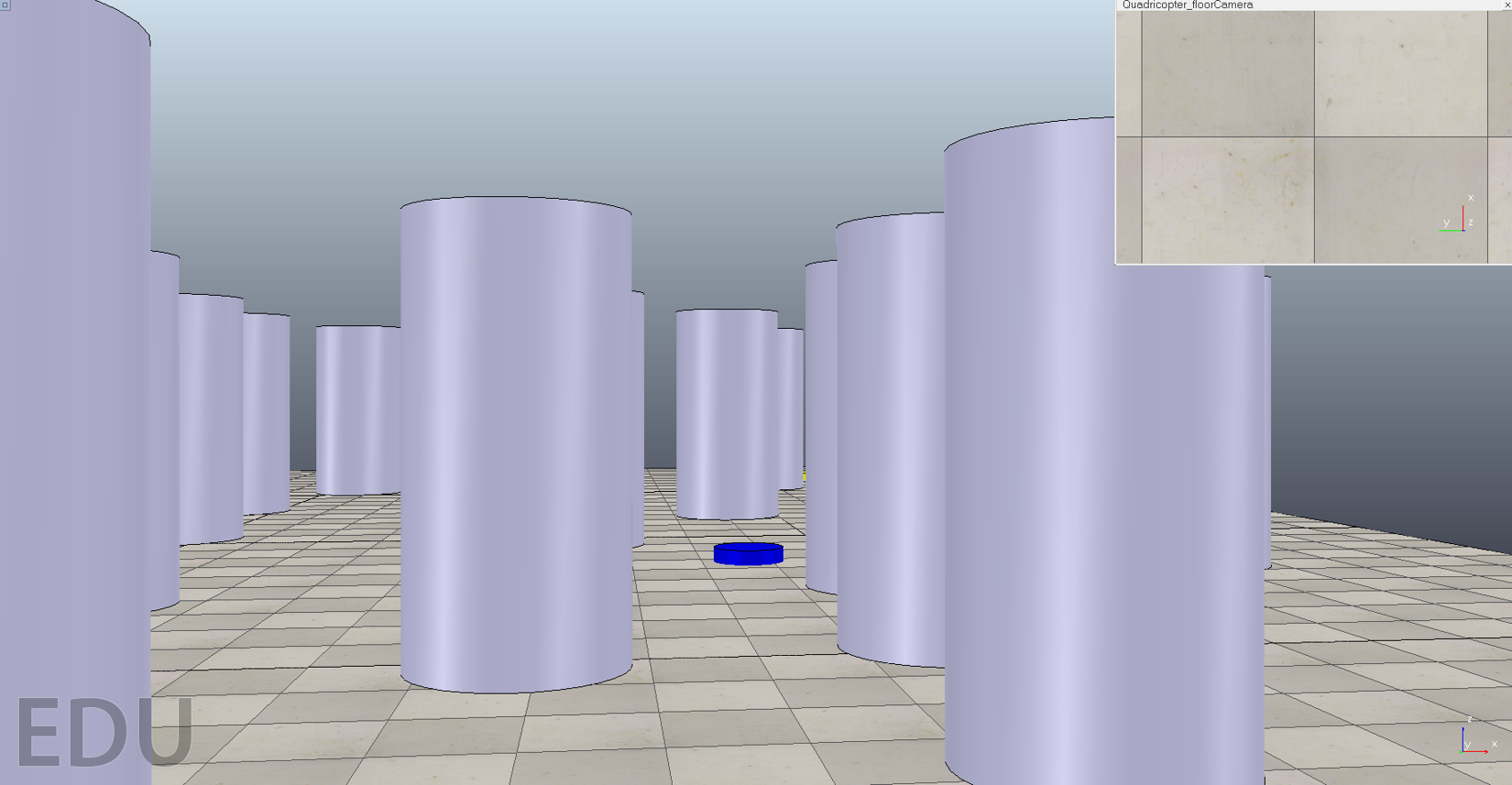}
  \caption{\small A first-person view that was provided to a human subject of the simulated environment.}
  \label{fig:scene}
  \vspace{-15pt}
\end{figure}

During the user study, each participant navigated a simulated quadrotor UAV in a forest-like environment, as shown in Fig. \ref{fig:setup}. The UAV and the environment were simulated using CoppeliaSim \cite{rohmer2013v}. The UAV had a radius of \unit[$0.25$]{m}. The UAV was equipped with a forward-facing camera (for navigation) and a bottom-facing camera (for ``inspection''). The view captured by the forward-facing camera was displayed on a 24-inch computer monitor, while the view captured by the bottom-facing camera was shown as a 5.5-inch insert at the top right of the screen, as shown in Fig.~\ref{fig:scene}.

The forest-like environment was a \unit[$25$]{m} by \unit[$15$]{m} rectangular area that contained 45 cylindrical tree-like obstacles, each with a radius of \unit[$0.5$]{m}. The environment also contained four targets. For each trial, the participant was asked to fly the UAV through the simulated environment to ``inspect'' each of the four targets. 

For this environment, the safety set is the set of all states where the robot is not in contact with an obstacle. Approximating the quadrotor as a cylindrical disc, then the safety set can be described by using the following $h(x)$ in~\eqref{eqn:safety_set}: 
\begin{equation}\label{eqn:barrier_function}
    h(x)= \norm{x_1-x_{1,o}}^2 - r^2,
\end{equation}
where $x_{1,o}$ denotes the position of the obstacle, $r$ denotes the sum of the radius of the UAV and the obstacle.

A 3D Systems Touch Haptic Device was used as the interface to control the motion of the UAV. The height of the robot above the floor was fixed and the user only had control of the robot's horizontal position and yaw angle. The user controlled the robot from a first-person perspective so that moving the stylus forward (i.e. away from the user) would result in a motion in the direction of the UAV's front-facing camera, moving the stylus to the left would result in the UAV banking left, and so on. The yaw angle of the UAV was controlled using the two buttons on the stylus of the haptic device, with one button commanding a counterclockwise rotation and the other commanding a clockwise rotation.
The stylus command was mapped to the UAV's commanded velocity $x_{2d}$ through a constant of \unit[$2$]{$\frac{m/s}{cm}$}, with a dead-zone of 1 cm to help the user hover the UAV in place. The rate of yaw rotation was \unit[$\frac{\pi}{4}$]{rad/s} when a button was pressed. A constant \unit[$0.3$]{N} restoring force was implemented to help the user return the haptic joystick to the deadzone. We note that these restoring forces were always present, and were simply added to any forces generated by the teleoperation conditions. The parameters were set up through several pilot trials to make the user feel comfortable while also keeping the implementation of our system feasible.

\subsection{Experimental Procedure}
Twelve subjects participated in this user study (aged $23$-$34$, two females, one left-handed). Participants held the stylus of the haptic device with their dominant hand, and the device was positioned to be in line with the subject’s corresponding shoulder.
Each subject completed four trials with each one of the conditions shown in Fig.~\ref{fig:paradigms} (for a total of 16 trials). 

To reduce learning, fatigue, and order effects, the presentation order of the control methods was counterbalanced using a Latin Square. To prevent the subject from memorizing the spatial layout of the task, the locations of the tree-like obstacles and the targets were randomly assigned before each trial. For all trials, the UAV started at the same location with the same camera orientation.

After providing informed consent, the subject was shown how to control the simulated UAV and was given the opportunity to practice flying in an environment without any obstacles. The subject practiced flying without using any assistance. Subjects were told that during some trials, they may feel forced from the 3D Systems Touch. They were given no other information about the four test conditions. Participants were told to fly the UAV in the simulated environment and ``inspect'' each one of the four targets by hovering the UAV over the target, and pressing a button on the keyboard to take a picture with the bottom-facing camera. The participants were asked to complete the task as quickly as possible without colliding with the simulated obstacles while prioritizing safety over speed.

Each trial started when the participant issued the first non-zero velocity command to the UAV.  A successful trial ended when the participant inspected all four targets. A failed trial ended when the participant crashed the UAV into the obstacle in such a way that the simulated UAV lost flight. 
Participants completed a block of four trials for each condition, then provided subjective measures of their experience using the NASA Task Load Index (NASA-TLX) \cite{TLX} and a custom survey. After the subject completed all four conditions, they completed a final survey where they rank-ordered the four conditions according to several criteria. Details of the two custom surveys are presented with the results.


\section{RESULTS}\label{sec:results}
\begin{figure}[t] 
  \centering
  \includegraphics[width=1\columnwidth, trim={0.6cm 0.5cm 0 0cm},clip]{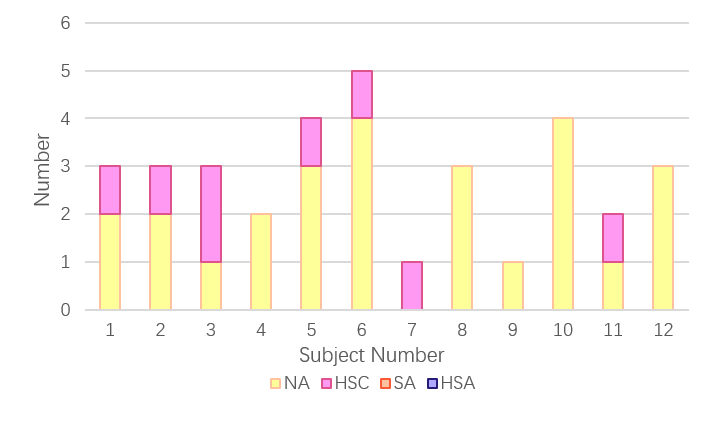}
  \vspace{-25pt}
  \caption{\small Results of the number of failures.}
    \vspace{-5pt}

  \label{fig:failures}
\end{figure}

We evaluate the results of the user study using both objective and subjective criteria.
\paragraph{Number of Failures}
We recorded the number of successful trials and failures. The numbers of failures are shown in Fig. \ref{fig:failures}. For total 48 trials of each condition, there were 8 failed trials under HSC and 26 failed trials under NA. As expected, no failures were resulted under HSA and SA, since they issued the safe control signal to the UAV. 
\paragraph{Task Performance Metrics}

\begin{figure}[t] 
\centering
\includegraphics[width=1\columnwidth,trim={0.4cm 0cm 0cm 0cm},clip]{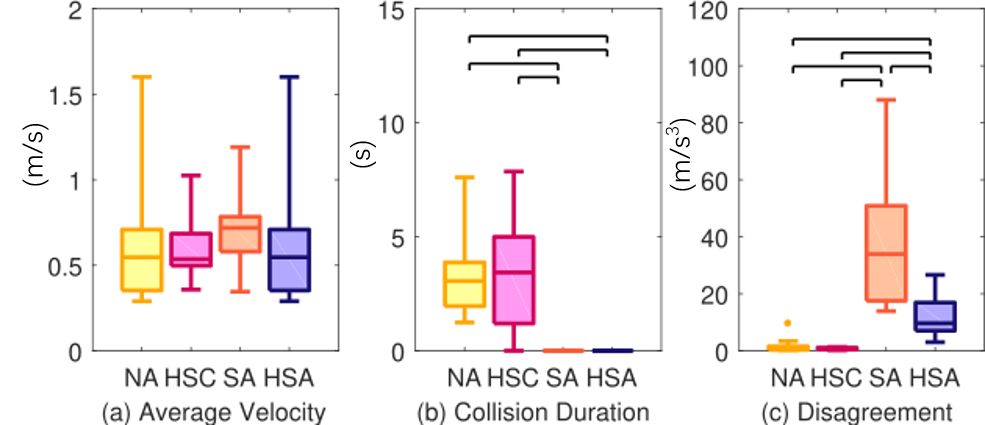}
\vspace{-10pt}
    \caption{\small Results of the metrics of the successful trials. Black brackets indicate significance between conditions at the p$<$0.05 level. A greater value of  $V_{avg}$ indicates a better performance. Lower values of $T_{collision}$ and $Disagreement$ are preferable.}
\label{fig:metrics}
\vspace{-15pt}
\end{figure}

For each successful trial, we computed the following metrics when using each of the four conditions:
\begin{itemize}
    \item  $V_{avg}$: The average linear velocity of the virtual UAV. 
    \item $T_{collision}$: It was possible for the UAV to lightly contact the obstacle and remain in stable flight. $T_{collision}$ is the total duration of time that the UAV was in contact with the obstacles. 
    \item $Disagreement$: The norm of the difference between the human operator's command and the command returned by the shared autonomy divided by trial time.  
\end{itemize}

Task metrics were analyzed using a repeated-measures analysis of the variance (rANOVA) to determine whether the condition used to control the UAV had any effect on task performance or user experience. When a significant difference in subject performance was found, Tukey's test was performed at a confidence level of $\alpha = 0.05$ to determine which methods led to significant differences in the metric.

As shown in Fig. \ref{fig:metrics}, there is a significant effect on the collision duration (F(3,27) = 6.7912, p = 0.001474). HSA and SA have shorter collision duration compared with HSC and NA. When considering the results of the disagreement, the four conditions also have a significant effect (F(3,27) = 22.136, p = 1.9255e-07). HSC and NA caused less disagreement as compared with HSA and SA. In addition, HSA resulted in significantly less disagreement as compared with SA.
The results did not show a significant difference among four tested conditions, in terms of average velocity (F(3,27) = 0.94655, p = 0.43199) for the successfully completed trials. 
\paragraph{NASA-TLX}

\begin{figure*}[t] 
\centering
\includegraphics[width=2\columnwidth,trim={0cm 0cm 0 0cm},clip]{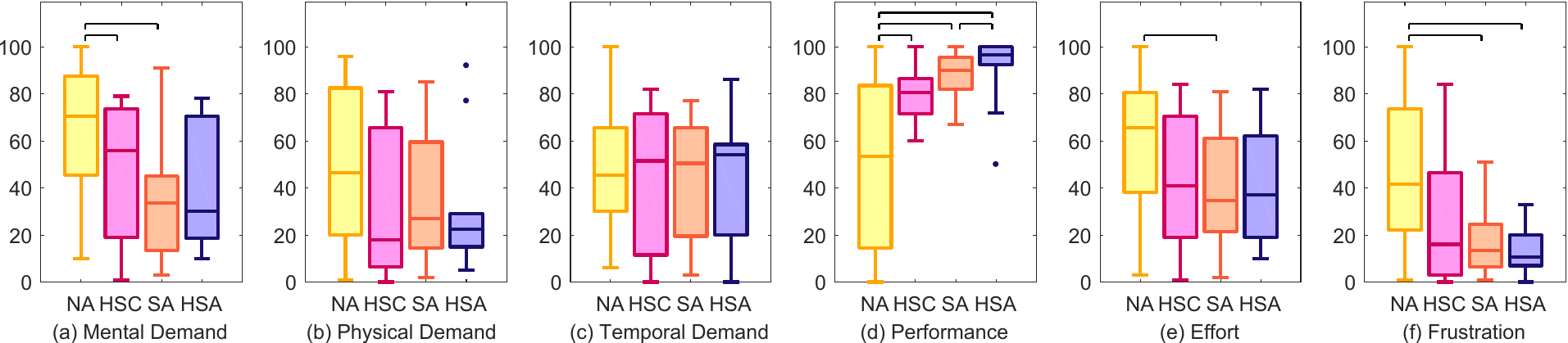}
\caption{\small  Results of the NASA-TLX survey. Black brackets indicate significance between conditions at the p$<$0.05 level.}
\label{fig:resultsTLX}
\end{figure*}
All users' subjective ratings of the task performance and workload are shown in Fig. \ref{fig:resultsTLX}. No significant differences were found when comparing conditions on ratings of physical demand (F(3,33) = 2.1183, p = 0.11671) and temporal demand (F(3,33)= 0.13142, p = 0.9407). Significant differences were found when comparing the effect of each condition on mental demand (F(3,33)= 6.2525, p = 0.0018), performance (F(3,33) = 15.048, p = 2.4171e-06), effort (F(3,33) = 5.582, P = 0.0033) and frustration (F(3,33)= 9.3405, p = 0.00013). HSC and SA have significantly lower mental demand compared with NA. Participants reported a higher level of performance for HSA as compared with HSC and NA. Subjects also thought they performed better with HSC and SA than with NA. In addition, NA led to higher levels of effort as compared to SA and led to higher levels of frustration to participants as compared with both HSA and SA.

\paragraph{Post-condition Surveys}

\begin{figure*}[t] 
\centering
\includegraphics[width=2\columnwidth,trim={0cm 0cm 0 0cm},clip]{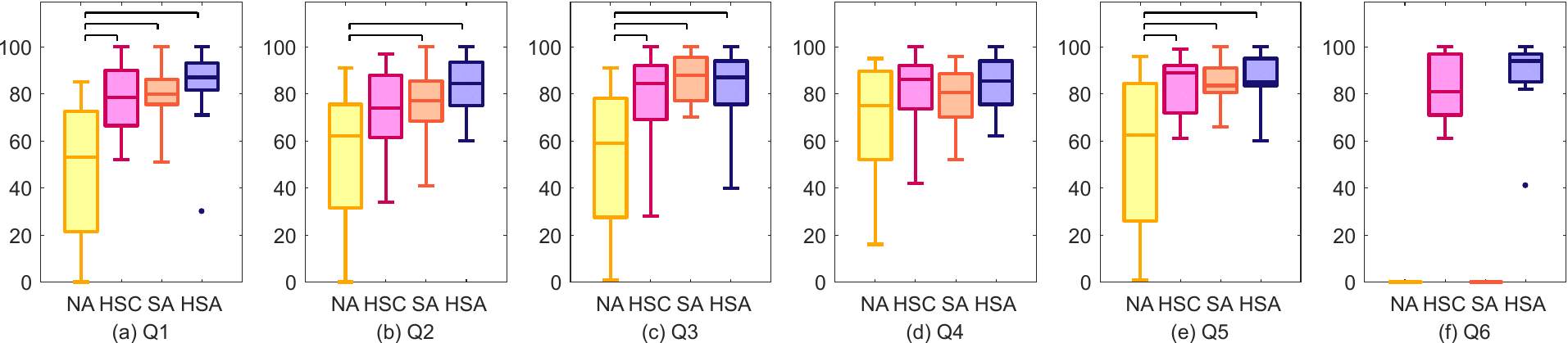}

\caption{\small  Results of the post-condition surveys. Black brackets indicate significance between conditions at the p$<$0.05 level.}
 \vspace{-10pt}
\label{fig:post_condition}
\end{figure*}

After completing the task with each condition, participants rated the condition on a scale that ranged from 0 to 100 using the following questions:
\begin{lenumerate}{Q}
\item How easy was it to complete the task?
\item How easy was it to control the robot?
\item How confident were you in your ability to move the robot?
\item How well did the robot’s motion match your intended motion?
\item How would you rate your overall experience with this condition?
\item If you felt haptic feedback, how much did the feedback help you navigate the robot?
\end{lenumerate}
The results of the surveys are shown in Fig. \ref{fig:post_condition}. Responses to Q1-Q5 were analyzed with rANOVA with Tukey's post-hoc test with $\alpha = 0.05$. The results showed that subjects generally rated the three assistive conditions more positively than NA, with significance being found for Q1 (F(3,33) = 13.055, p = 0.0001198), Q2 (F(3,33) = 7.2053, p = 0.00075369), Q3 (F(3,33) = 10.618, p = 4.8663e-05), and Q5 (F(3,33) = 10.674, p = 4.6715e-05). Furthermore, responses to Q6 indicate that subjects felt that the haptic feedback helped them to navigate the robot in the HSA and HSC conditions. 

\paragraph{Post-experiment Rankings}

\begin{figure}[t] 
\centering
\includegraphics[width=0.95\columnwidth,trim={0cm 0cm 0 0cm},clip]{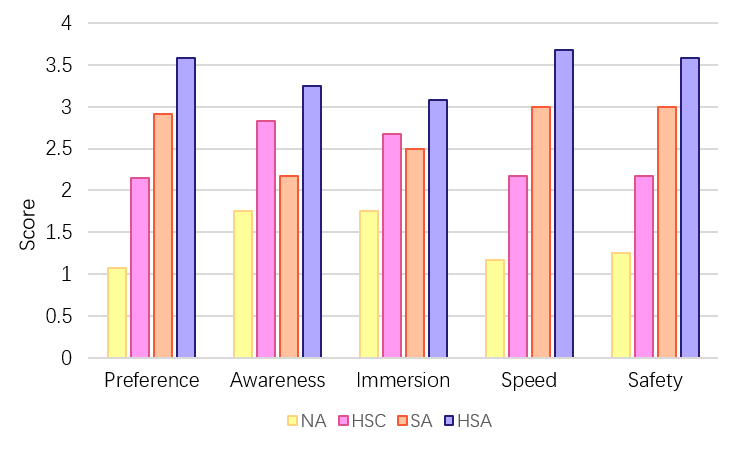}
\vspace{-10pt}
\caption{\small  Results of the post-experiment ranking surveys with respect to five questions.}
\vspace{-25pt}
\label{fig:Rankings}
\end{figure}

After all four conditions were tested, each subject completed a final survey in which they ranked the four conditions according to their favorite method, the method that gave them the best awareness of the robot's environment, the method that made them feel most immersed in the robot's environment, the method they would choose to accomplish a task quickly, and the method they would choose to accomplish a task safely.

We assigned a score of 4 to the best-ranked method and a score of 1 to the worst-ranked method. We then found the average score of all the subjects' ranking for each condition. Fig.~\ref{fig:Rankings} shows the average ranking for each condition. HSA is the most preferred condition according to each of the metrics. Overall, subjects indicated that SA was their second favorite method overall and was their second preferred method to complete a task quickly and safely. Subjects indicated that HSC resulted in the second highest levels of awareness of and immersion in the robot's environment.

\section{DISCUSSION}\label{sec:discussion}

\textit{Hypothesis} \ref{H2} is directly supported by the results of the rankings that users picked HSA and HSC as the conditions that made them most aware of and feel most immersed in the robot's environment. HSC teleoperation resulted in fewer failed trials and lower user ratings of mental demand than NA, further supporting \ref{H2}.


Our results also support \textit{Hypothesis} \ref{H3}. Because the CBF-based shared autonomy guarantees the safety of the UAV, no subject collided with the UAV when using either HSA or SA. Consistent with this fact, subjects also indicated that HSA and SA would be their top two choices to complete a task safely. No significant differences were found in the average velocity of the UAV between the conditions. However, subjects indicated that HSA and SA would be their top two choices to complete a task quickly, so subjects may feel most confident when flying the UAV with these conditions, which could result in faster task performance.

 

During this task, the user's issues commands were significantly closer to the safe commands when using HSA as compared to SA, confirming \textit{Hypothesis} \ref{H4}. 
We note that this metric is only calculated for successful trials. Therefore, it makes sense that conditions without any shared autonomy (NA and HSC) resulted in the lowest levels of human-robot disagreement because the UAV would crash if the user's control signal is too far from a safe control signal, resulting in a failed trial. Both SA and HSA have significantly higher levels of human-robot disagreement than NA and HSC. However, the results show that adding haptic feedback to a shared autonomy paradigm can lead the human to issue commands that are much closer to the commands issued by the autonomous assistance. 


Because \textit{Hypothesis} \ref{H2}-\ref{H4} are all supported by the user study results, we note the providing any form of assistance via HSC, SA, or HSA can improve the user's performance and satisfaction when flying a UAV. HSA consistently outperforms the other assistive conditions, SA and HSC, as measured by user performance and user ratings. Therefore, our findings confirm \textit{Hypothesis} \ref{H1}.


\section{CONCLUSIONS AND FUTURE WORK}\label{sec:conclusion}
In this paper, we proposed a novel haptic shared autonomy teleoperation paradigm that uses haptic feedback to inform the user about the inner state of a shared autonomy paradigm. Shared autonomy teleoperation can guarantee safety, but does so by reducing the human operator's level of control authority. This can lead to reduced levels of human-robot agreement and user satisfaction. Importantly, we find that adding haptic feedback to shared-autonomy teleoperation can improve both human-robot agreement and user satisfaction in shared autonomy teleoperation, while still providing formal guarantees to improve upon human-level performance. 

Our immediate future work includes testing haptic shared autonomy on a real UAV. In this study, the haptic shared autonomy system never presented any stability issues. However, given that we are using force feedback, in the future we will update our control methods to guarantee system stability.



\bibliographystyle{IEEEtran} 
\bibliography{paper}
\end{document}